\documentclass[12pt]{article}
\usepackage[noblocks]{authblk}

\usepackage{natbib, graphicx, caption, subcaption}
\usepackage[letterpaper, margin=2cm]{geometry}

\date{}
\usepackage[compact]{titlesec}
\titlespacing{\section}{0pt}{-1.0ex}{-1.0ex}
\titlespacing{\subsection}{0pt}{1ex}{0ex}
\titlespacing{\subsubsection}{0pt}{0.5ex}{0ex}

\graphicspath{ {./Fig.s/} }

\begin{document}

\title{\normalsize\textbf {Spatially Aware Deep Learning for Microclimate Prediction from High-Resolution Geospatial Imagery}}

\author[1]{\normalsize Idan Sulami}
\author[1]{\normalsize Alon Itzkovitch}
\author[2]{\normalsize Michael R. Kearney}
\author[3]{\normalsize Moni Shahar}
\author[1,*]{\normalsize Ofir Levy}

\affil[1]{\normalsize Tel Aviv University, Faculty of Life Sciences, School of Zoology, Tel Aviv, Israel}
\affil[2]{\normalsize School of BioSciences, The University of Melbourne, Melbourne, Victoria, Australia}
\affil[3]{\normalsize Tel Aviv University, Center for Artificial Intelligence \& Data Science, Tel Aviv, Israel}
\maketitle

\noindent \textbf{*Corresponding author:} Ofir Levy; mailing address: School of Zoology, Faculty of Life Sciences, Tel Aviv University, 6997801; phone: +972504384717; fax: +97236409403; email: levyofir@tauex.tau.ac.il.

\noindent \textbf{Emails of co-authors:} Idan Sulami: idan.atj@gmail.com, Alon Itzkovitch: alonitzko@gmail.com, Michael R. Kearney: m.kearney@unimelb.edu.au, Moni Shahar: monishahar@tauex.tau.ac.il.

\noindent \textbf{Data accessibility statement:} Should the manuscript be accepted, the data and code with full equations will be archived in an appropriate public repository (e.g., Dryad, Figshare, or Zenodo) and the data DOI will be included at the end of the article. For review purposes, the code and sample data can be found in https://github.com/levyofi/Sulami\_et\_al\_2026. 

\noindent \textbf{Authors' contributions:} All authors conceived and designed the study, wrote the first draft and participated in revising the manuscript. OL collected the data and contributed mentorship. IS and AI contributed modeling and validation. 

\section*{ABSTRACT}
\noindent
Microclimate models are essential for linking climate to ecological processes, yet most physically based frameworks estimate temperature independently for each spatial unit and rely on simplified representations of lateral heat exchange. As a result, the spatial scales over which surrounding environmental conditions influence local microclimates remain poorly quantified, particularly in heterogeneous landscapes. Here, we show how remote sensing can help quantify the contribution of spatial context to microclimate temperature predictions. Building on convolutional neural network principles, we designed a task-specific deep neural network and trained a series of models in which the spatial extent of input data was systematically varied. Drone-derived spatial layers and meteorological data were used to predict ground temperature at a focal location, allowing direct assessment of how prediction accuracy changes with increasing spatial context. Our results show that incorporating spatially adjacent information substantially improves prediction accuracy, with diminishing returns beyond spatial extents of approximately 5--7~m. This characteristic scale indicates that ground temperatures are influenced not only by local surface properties, but also by horizontal heat transfer and radiative interactions operating across neighboring microhabitats. The magnitude of spatial effects varied systematically with time of day, microhabitat type, and local environmental characteristics, highlighting context-dependent spatial coupling in microclimate formation. By treating deep learning as a diagnostic tool rather than solely a predictive one, our approach provides a general and transferable method for quantifying spatial dependencies in microclimate models and informing the development of hybrid mechanistic--data-driven approaches that explicitly account for spatial interactions while retaining physical interpretability.

\noindent \textbf{Keywords:} landscape, modeling, drone, climate, neural networks, management, rocks, vegetation.

\section{INTRODUCTION}

\noindent Climate conditions are fundamentally important in shaping how organisms experience the environment, from shaping long evolutionary processes to momentary decisions, and from ecological communities to individuals \citep{potter_2013, pincebourde_2020}. However, studying the impacts of climate on organisms is tricky since individuals experience the climate conditions in their proximity (meters to cm) while most of our measurements and online climate data are available at much larger scales, often more than 10 km. Thus, to investigate the direct links between climate and the physiology, behaviour, distribution and abundance of organisms, fine spatial and temporal resolution climate data are required (Kearney et al. 2012). To bridge the gap between the scales of climate data and organismal experience, ecologists have been developing microclimate models that translate climate conditions into microclimates \citep{kearney_2017, maclean_2019}. Using mechanistic approaches grounded by physical heat-transfer rules, these models calculate the heat-balance between the ground, the soil, and the atmosphere near the ground at any particular location. However, making accurate predictions of natural phenomena is one of the big challenges in science, and microclimate predictions often suffer from errors and biases. 

Like nearly all models, the accuracy of microclimate models can suffer from inaccuracies in parameterization of physical characteristics controlling heat transfer or from simplifications embedded within the models. Two primary challenges persist: poor parameterization and oversimplified assumptions. One major simplification of current physical models is that they frequently overlook the influence of nearby microhabitats. For instance, objects near modeled locations may reflect solar radiation \citep{kearney_2017}, and adjacent vegetation can cool air temperatures \citep{rahman_2020}. Although advanced remote sensing techniques such as drones provide detailed terrain, vegetation structure, and contextual maps \citep{duffy_2021}, it remains unclear how effectively such high-resolution spatial data can improve model predictions, and whether it is important to incorporate nearby information in our models.

Deep learning is well-suited for capturing nonlinear spatial dependencies in complex environmental data \citep{lecun_2015}, particularly when high-resolution spatial inputs from drone-based imagery are combined with meteorological measurements \citep{campsvalls_2025, schneider_2023, slater_2023, levy_2024}. Here, we developed a task-specific deep neural network designed to quantify the spatial extent over which nearby environmental information contributes to microclimate predictions. Although our architecture builds on convolutional neural network principles commonly used in computer vision, it was explicitly adapted and restructured for microclimate modeling by incorporating multiple environmental feature layers, continuous temperature prediction, and a controlled manipulation of spatial input size. Using this framework, we trained a series of models with systematically increasing spatial context to evaluate how spatial information influences ground temperature estimates at a focal location. This design allows spatial dependencies to be assessed directly, identifies characteristic spatial scales of influence, and reveals how these scales vary across times of day, microhabitats, and spatial features. By developing and applying a dedicated deep learning architecture as an exploratory modeling tool, our study provides a generalizable methodological approach for probing spatial structure in microclimate models and for informing the development of hybrid mechanistic–data-driven frameworks.

\section{MATERIALS AND METHODS}
\subsection{General approach}
\noindent 
We developed a deep learning–based methodological framework to quantify the contribution of spatial context to microclimate temperature predictions at a focal location. Spatial information surrounding each prediction point was parameterized using high-resolution drone imagery collected during flight mapping missions. Using a task-specific convolutional neural network, we systematically varied the spatial extent of input data to examine how increasing spatial context influences prediction accuracy. This design allowed us to identify characteristic spatial scales at which nearby information improves model performance and to evaluate how these scales vary across microhabitats, times of day, and spatial features that are relevant to microclimate formation.

\begin{figure}
    \centering
    \begin{minipage}{0.5\textwidth}
        \includegraphics[width=\textwidth]{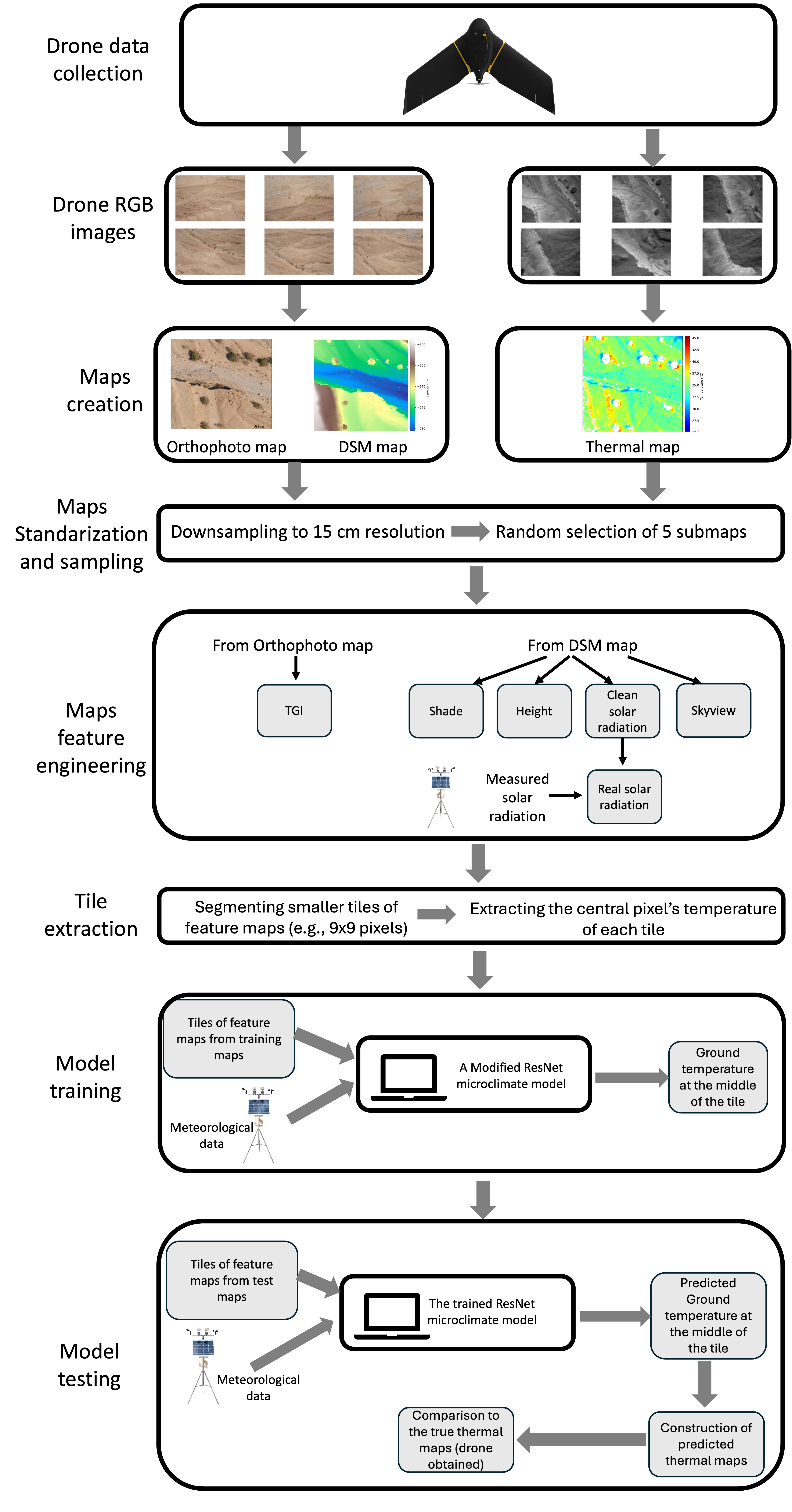}
    \end{minipage}%
    \hfill
    \begin{minipage}{0.45\textwidth}
        \caption{Flowchart of the data and modeling pipeline used in our study. Drone mapping missions were conducted to simultaneously collect RGB and thermal imagery, which were processed into orthophoto, Digital Surface Model (DSM), and thermal maps. All maps were standardized to a 15 cm² resolution, and five submaps were randomly sampled from each. From these, five feature maps were generated—Triangular Greenness Index (TGI), shade, height, skyview, and solar radiation—and segmented into small tiles. Each tile was paired with the temperature at its central pixel to train a modified ResNet model. Separate models were trained for different tile sizes to evaluate how spatial context influences model performance. Model accuracy was assessed across varying microhabitats, times of day, and environmental conditions. }
        \label{fig:complete_data_flow}
    \end{minipage}
\end{figure}

\subsection{Drone-based Deep Learning Microclimate Model}

\subsubsection{Field Data Collection}
\noindent We conducted field data collection to measure various meteorological features (i.e., model predictors), both constant and spatially variable, needed to train and test the deep learning model. These features were measured either in situ using a drone and a meteorological station, or sourced from the Global Land Data Assimilation System (GLDAS, \citealt{TheGlobalLandDataAssimilationSystem}). Most features remained constant across each map, except for five spatially varying factors: solar radiation, shade, skyview, vegetation, and height above the ground (Table S1). Our data collection took place between June 2019 and May 2021, when we completed 33 drone flights across eight separate days, capturing morning-to-evening variability (Table S2).

\paragraph*{\textit{Study Site}} Field data collection took place in the Judean Desert, Israel (31°28'N, 35°10'E), adjacent to the Dead Sea (~400 m below sea level). This region primarily consists of sparse perennial shrubs and annual grasses \citep{moncaz_2012}. During summer, ground temperatures in exposed areas range from approximately 30°C early in the morning to around 44°C midday, while shaded areas under rocks experience maximum temperatures around 37°C \citep{levy_2016a}. In winter, open ground temperatures typically reach 27°C, whereas shaded areas maintain slightly cooler temperatures around 25°C \citep{levy_2016a}. Our primary study location was Parking Tse'elim River (31°21'04.8"N 35°21'11"E), characterized by rocky terrain and minimal vegetation.

\paragraph*{\textit{Meteorological Data Collection}}
At the start of each field session, we installed a portable meteorological station (MaxiMet GMX501, GILL Instruments, UK) in a flat, fully exposed area within the mapping zone. This station recorded solar radiation, air and ground temperatures, wind speed and direction, relative humidity, and air pressure at ten-minute intervals.

\paragraph*{\textit{Online Meteorological Data}}
We complemented field-collected data with additional climate variables obtained from the GLDAS database, including critical parameters such as albedo and soil moisture, essential for accurate microclimate modeling (Table S1).

\paragraph*{\textit{Drone Imagery}}
We collected spatial information using an eBee X mapping drone (SenseFly, Switzerland) equipped with a Duet T dual RGB and thermal camera (±5°C accuracy; see Map Development for calibration). This dual-lens camera collected data to serve as features to the deep learning model, based on the RGB camera, and the expected model output, the ground temperatures, based on the thermal camera. Surveys were strategically conducted to capture diverse daily and seasonal conditions (Table S2), covering areas from approximately 0.3 km² to 1 km² with a standard 65\% image overlap. Flight missions were planned and executed using eMotion 3 flight-planning software (SenseFly, Switzerland), lasting between 20 to 60 minutes. Images captured during flights were stored on SD cards and transferred to a laptop post-flight for map development. 

\subsubsection{\textit{Spatial Tiles Map Development}}
Drone imagery and meteorological data underwent a two-stage processing pipeline to generate the spatial features for the model. Initially, drone images were processed through photogrammetry software (Pix4DMapper v4.23, Pix4D Inc., USA), yielding high-resolution RGB orthophotos, Digital Surface Models (DSMs), and Digital Terrain Models (DTMs) at a 3 cm\textsuperscript{2} spatial resolution. Thermal imagery produced temperature validation maps at a 15 cm\textsuperscript{2} resolution, which we calibrated using temperature measurements from our mobile weather station to account for sensor accuracy (±5°C), assuming a constant calibration value across all pixels. Next, all spatial datasets were resampled to match the 15 cm resolution of thermal maps, after which five randomly selected 1024×1024 pixel maps per flight were used for further analysis. Despite GPS positional accuracy of approximately 2 m, within-flight images were perfectly aligned.

In the second stage, we used the DSM, DTM, and RGB orthophoto maps data to create the five spatial features: solar radiation (W m\textsuperscript{–2}), shade (presence/absence), skyview (\%), height (m), and vegetation cover \citep[Triangular Greenness Index, TGI;][]{star_2020}. Such features are regularly used to parameterize traditional physically-based microclimate models \citep{kearney_2020, maclean_2019}. These layers were generated using R scripts and GRASS GIS commands \citep[][ver. 7.4.0]{GRASS_GIS_software}, integrated within R \citep[][ver. 4.1.0]{R_Core_Team}, facilitated by the \textit{rgrass7} package \citep{BIVAND20001043}. Both stages of the mapping process are illustrated in Fig. \ref{fig:complete_data_flow}, with detailed methodological descriptions available in the supplementary material.

\subsection{Model Dataset Preparation}

\noindent After generating the initial set of feature maps (as described above), we trimmed their edges to minimize inaccuracies where information in nearby pixels is necessary —particularly for skyview calculations. This resulted in consistent maps of dimensions 1000$\times$1000 pixels. We then stacked the maps to create five-channel composite images (i.e., a matrix of 5$\times$1000$\times$1000). Before training, we standardized the maps and the corresponding meteorological variables to a common scale by subtracting their mean and dividing by their standard deviation. This common preprocessing step improves model convergence and predictive accuracy. We also centered the ground temperature labels (i.e., the values that the model should predict). After standardization, each full-sized (1000$\times$1000 pixels) five-channel map underwent partitioning into smaller spatial tiles to facilitate model training and explore the role of spatial information around the prediction target. 

\begin{figure}
    \centering
    \includegraphics[width=1\textwidth]{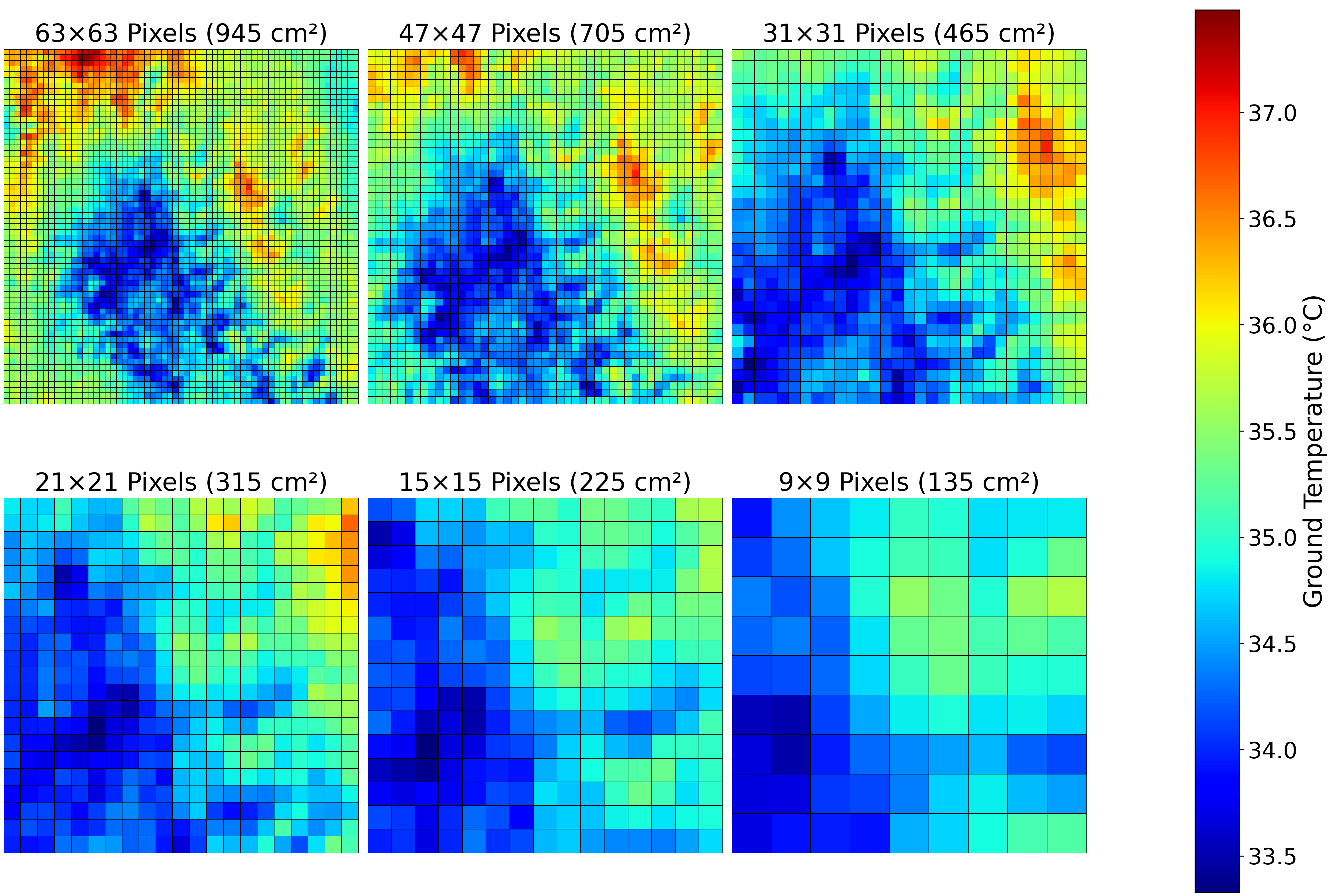}
    \caption{Illustration of tiles at different spatial scales used in the study. Each panel shows a tile of a specific size, ranging from 9×9 to 63×63 pixels. Larger tiles retain broader spatial context and environmental gradients, while smaller tiles capture more localized information with less surrounding context. Each panel displays ground-truth thermal maps.}
    \label{fig:tiles}
\end{figure}

To analyze how spatial context affects temperature predictions, we trained seven models corresponding to seven sizes of tiles: 9$\times$9 (1.35 m\textsuperscript{2}), 15$\times$15 (2.25 m\textsuperscript{2}), 21$\times$21 (3.15 m\textsuperscript{2}), 31$\times$31 (4.65 m\textsuperscript{2}), 47$\times$47 (7.05 m\textsuperscript{2}), 63$\times$63 (9.45 m\textsuperscript{2}), and 81$\times$81 (12.15 m\textsuperscript{2}) pixels (Fig. \ref{fig:tiles}). Specifically, we employed a sliding-window approach (i.e., convolutional filter), moving horizontally and vertically across each map with an 11-pixel stride, to generate smaller spatial sub-maps, or tiles. Each sub-map was labeled with the temperature of its central pixel, enabling an assessment of how spatial context influenced prediction accuracy. These smaller five-channel sub-maps, combined with their corresponding meteorological data, formed the input dataset for our deep learning model. Figure \ref{fig:complete_data_flow} summarizes this complete dataset processing pipeline.

\subsection{Deep Learning Model Architecture}

\noindent To predict microclimate temperatures, we used a modified version of a deep neural network called ResNet-18 \citep{liang_2020}. This model is commonly used for image analysis because it efficiently recognizes complex patterns in spatial data \citep{tabak_2018}. Although initially designed for classifying images, its structure, which helps the network learn effectively from nearby data, also makes it useful for predicting continuous values like temperature. We adjusted the ResNet-18 model in two main ways to predict ground temperatures:

\begin{enumerate}
    \item \textbf{Input adjustments (additional information layers):}  
    Typically, the original ResNet-18 model analyzes images composed of three layers (or "channels"), corresponding to the colors red, green, and blue. In our case, the drone-based maps contained five layers of information (solar radiation, shade, skyview, vegetation, and height above the ground). Therefore, we adjusted the first part of the model to directly accept these five layers instead of the usual three.

    \item \textbf{Output adjustments (prediction layer):}  
    The original ResNet-18 was designed to classify images into distinct categories, such as identifying whether an image shows a cat or a dog. To better adapt the model for predicting temperatures, we adjusted its final layers. Specifically, we removed the original layers designed for classification tasks and replaced them with layers suited for numerical predictions. First, we added a layer that simplifies the complex patterns learned by the model from 512 (tile size <= 31$\times$31), 2048 (tile size 47$\times$47), or 5096 (tile size 67$\times$67 and 81$\times$81) features down to 128 features, using a nonlinear function to enhance learning. Finally, we included a single-output layer that directly predicts temperature as a continuous value.

\end{enumerate}

\begin{figure}
    \centering
    \includegraphics[width=1\textwidth]{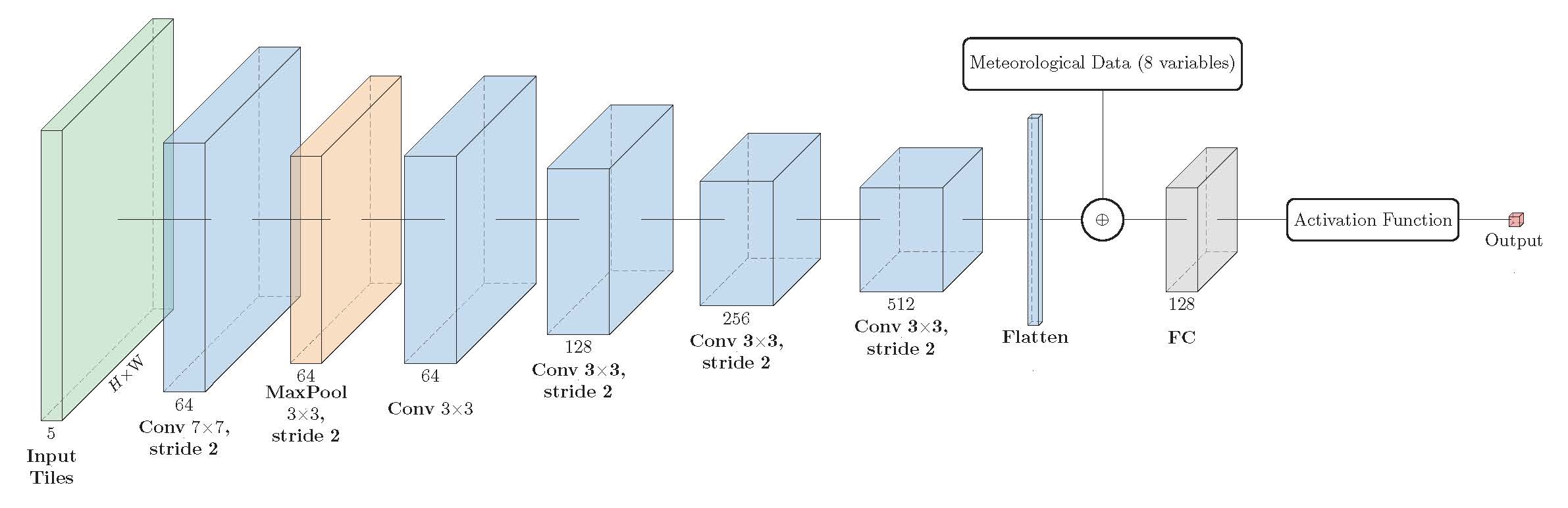}
    \caption{Convolutional neural network architecture for microclimate prediction. Five spatial input tiles are processed through successive convolutional and pooling layers, flattened, and concatenated with eight meteorological variables. The combined features are then passed through a fully connected layer and a nonlinear activation function to produce the final prediction, representing the temperature at the center of the tiles. Colors: green – input tiles; blue – convolutional layers; orange – pooling layers; grey – fully connected layer; red – predicted temperature.
}
    \label{fig:architecture}
\end{figure}

\noindent Figure \ref{fig:architecture} visually shows these modifications. We built and trained our adapted model using PyTorch \citep{NEURIPS2019_9015}, a software tool that simplifies creating and training neural networks.

\subsection{Training and Evaluation} 

\subsubsection*{Model Training}
\noindent We trained the model using data from 25 flights (conducted over five distinct field days) using 80\%/20\% of the tiles as the training and validating sets, respectively. We allowed all weights in the model to be unfrozen during training. The optimization of weights utilized the Adam algorithm with a learning rate set to 0.0001 \citep{kingma2015adam}. The Mean Squared Error (MSE) was selected as the loss function for the training process. For each tile size, we run the training for 100 epochs, saved the model at the end of each epoch, and chose the model with the lowest validation loss for further analysis.

For testing, we used the remaining 12 flights (from three different days). For these flights, we compared the prediction errors of the model at various tiles' sizes and microhabitats. To do so, we calculated the mean squared error (MSE):
\begin{equation} MSE = \frac{\sum_{i=1}^{N_{pixels}}e_i^2}{N_{pixels}}, \end{equation} 

\noindent where $e_i$ represents the model error at pixel \textit{i}. For each model size of tiles, we calculated each of these metrics separately for open and shaded microhabitats for the predictions of the test maps. Since our microclimate model calculates temperature for bare ground, we excluded pixels with vegetation cover (TGI > 0.04) from our analysis.

\subsection{Statistical Analysis}
\noindent We conducted two types of analyses to test our hypotheses.

In the first analysis, we examined the importance of spatial information and how it varies across microhabitats (open vs. shade) and dayparts (morning: 6:00–7:00, midday: 8:00–16:00, evening: 17:00–18:00). These microhabitats and dayparts reflect key environmental drivers of ground temperature, such as reduced solar radiation in shaded areas and rapid changes in solar angle during morning and evening hours. To test these effects, we fitted a Generalized Additive Mixed-Effects Model (GAMM) with the mean squared error (MSE) of each test map as the response variable. Tile size was included as a continuous predictor, and microhabitat and daypart as categorical predictors. To account for repeated measures from the same map, we included map identity as a random effect.


In the second analysis, we assessed how the importance of spatial information varies with different spatial features. Specifically, we selected 20 evenly spaced coordinates from each test map using the spatSample function from the terra R package \citep{terra_2025}. At each point, we extracted values for solar radiation, sky view factor, and TGI (representing shortwave radiation, longwave radiation, and vegetation, respectively). We also calculated the standard deviation of slope within a 2-meter radius to quantify local surface variation. For each deep learning model (per tile size), we extracted the squared error (SE) at each point. We then fitted GAMMs with SE as the response variable, tile size and each spatial feature as continuous predictors. To account for spatial autocorrelation, we included a smooth term for coordinates, and again treated map identity as a random effect.

All models were fitted using a Gamma distribution with a log link function. We used Akaike Information Criterion (AIC) \citep{burnham2002model} to compare model structures for how tile size interacts with the other explanatory variables. For categorical variables (microhabitat and daypart), we compared models in which tile size was modeled as either a separate smooth for each category or as a shared smooth with different intercepts. For continuous variables (e.g., solar radiation, sky view, TGI, and slope variability), we compared models where tile size and the feature were modeled as either joint or separate smooths. AIC comparisons are not shown here but are available in the supplementary code. All models were fitted using the bam function from the mgcv R package \citep{mgcv_2011, R_Core_Team}.

\section{RESULTS}

\noindent In general, model accuracy improved significantly as the training tile size increased. In particular, the mean squared error (MSE) decreased from 16.07 ± 11.59 (mean ± SD, $^{\circ}\mathrm{C}^2$) at a tile size of 5 to 1.58 ± 1.11 $^{\circ}\mathrm{C}^2$ at a tile size of 33. Beyond this tile size, mapping performance stabilized, indicating that a tile size of 33 pixels optimally captures relevant spatial patterns. 

\begin{figure}
    \centering
    \includegraphics[width=0.6\textwidth]{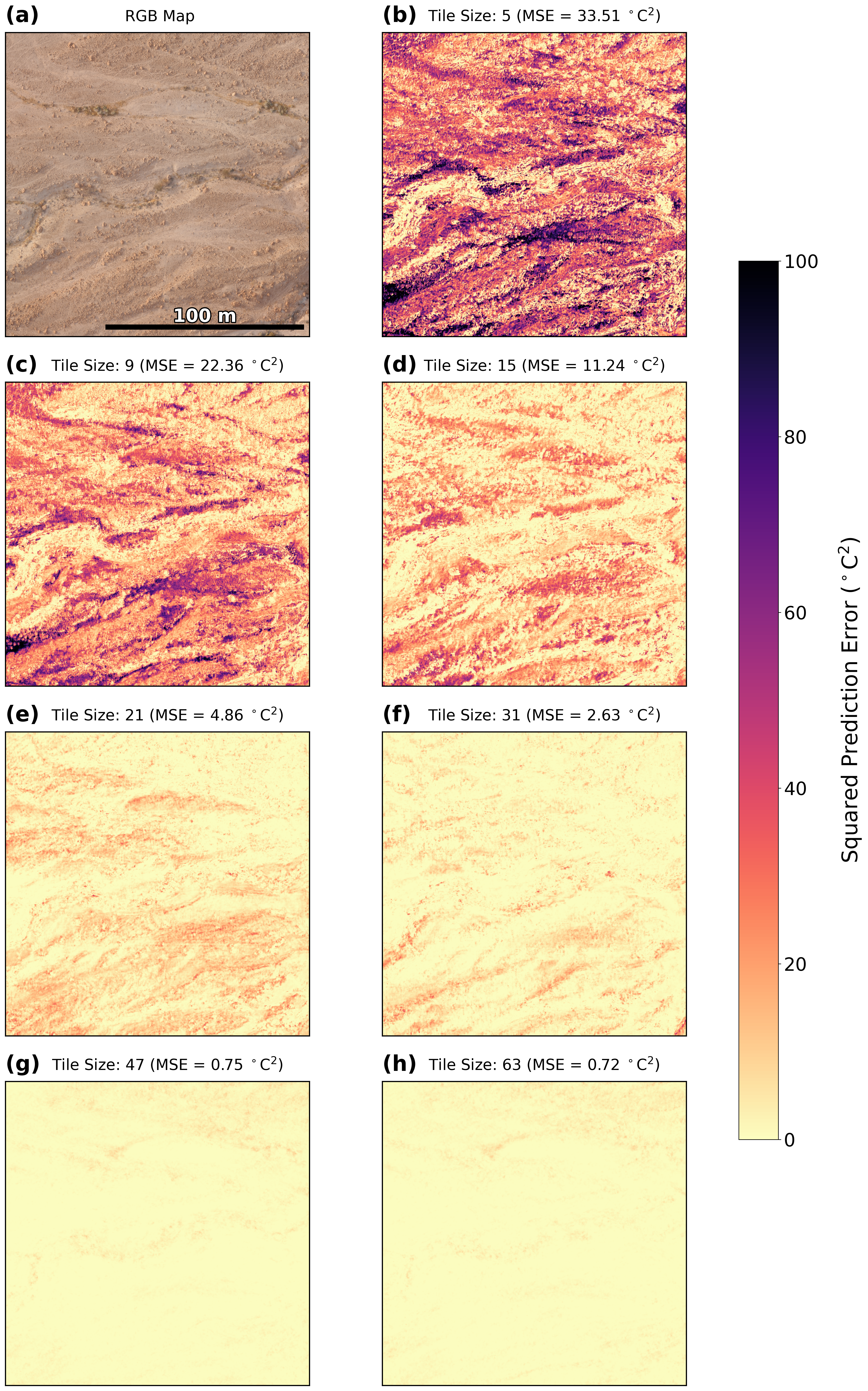}
    \caption{Progression of squared prediction error ($^\circ$C$^2$) with increasing tile size. (a) -- the reference RGB map. (b) through (h) -- the spatial distribution of squared prediction errors for models utilizing tile sizes ranging from $5\times5$ to $63\times63$ pixels, respectively. 
}
    \label{fig:mse_maps}
\end{figure}

\subsection{The importance of spatial information across microhabitats and daypart}

\begin{figure}
    \centering
    \includegraphics[width=1\textwidth]{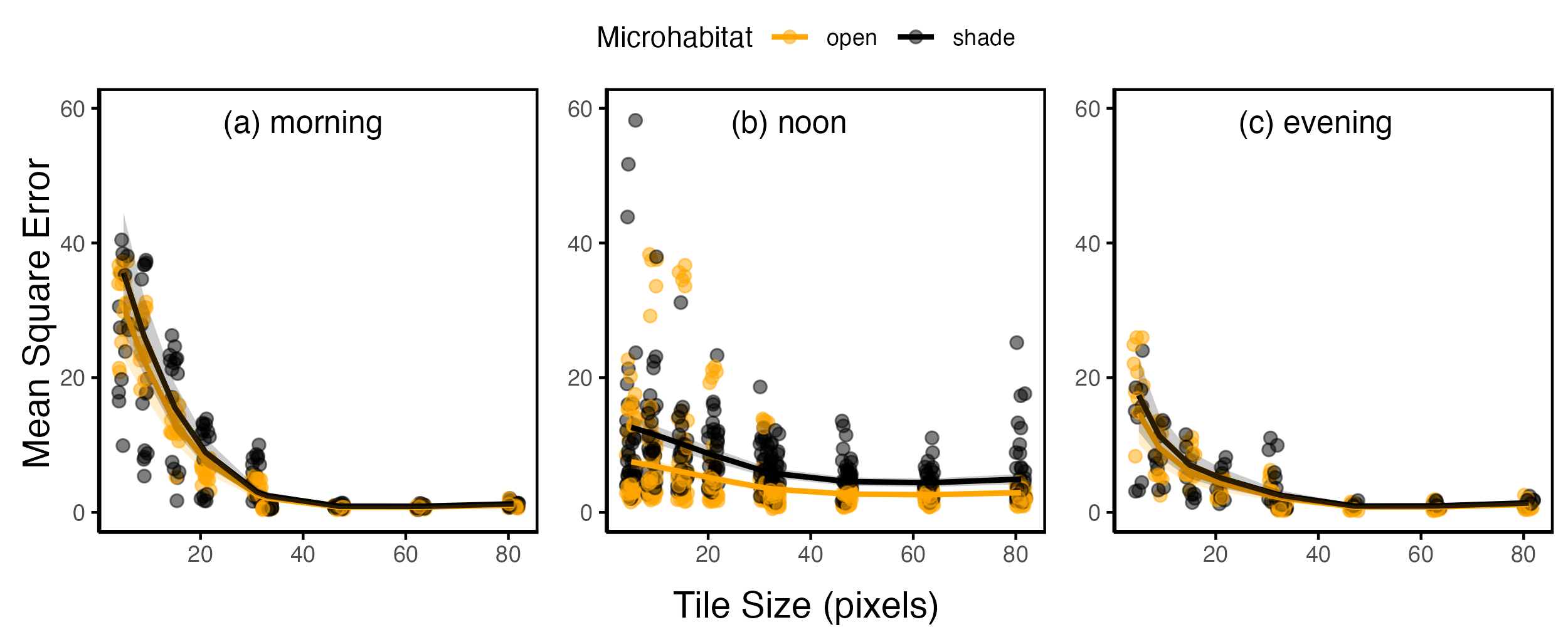}
    \caption{Model performance improved with increased spatial resolution, highlighting the value of spatial context in predicting ground temperature. Each panel shows how tile size affects prediction error (mean square error) across three dayparts: (a) morning, (b) noon, and (c) evening. Colored points represent predictions in open (orange) and shaded (black) microhabitats, and lines show smooth GAM fits with shaded confidence intervals.}  \label{fig:tile_size_vs_microhabitat_daypart}
\end{figure}

\noindent Our statistical analysis showed that the improvements in model performance with tile size varied across dayparts and microhabitats. Specifically, increasing tile size reduced prediction errors in all dayparts, with especially pronounced improvements in the morning and evening (\textit{Morning}: edf = 4.83, F = 105.01, \textit{p} < 0.001; \textit{Midday}: edf = 3.90, F = 66.58, \textit{p} < 0.001; \textit{Evening}: edf = 4.86, F = 289.89, \textit{p} < 0.001) (Fig.~\ref{fig:tile_size_vs_microhabitat_daypart}). Error reduction followed a decelerating trend, leveling off around tile sizes of $\sim$40 pixels ($\sim$6~m).

The parametric terms of our analysis revealed that prediction errors (SEs) were consistently higher in shaded microhabitats, though the magnitude of this difference varied by time of day. SEs were higher by $0.15 \pm 0.07~^{\circ}\mathrm{C}^2$ in shade during morning and evening (\textit{main effect of shade}: $t = 2.27$, \textit{p} = 0.024; \textit{shade} $\times$ \textit{morning interaction}: $0.02 \pm 0.11~^{\circ}\mathrm{C}^2$, $t = 0.15$, \textit{p} = 0.88), and substantially higher during midday (\textit{shade} $\times$ \textit{midday interaction}: $0.37 \pm 0.08~^{\circ}\mathrm{C}^2$, $t = 4.55$, \textit{p} < 0.001).

\subsection{The importance of spatial information across spatial features}
\begin{figure}
    \centering
    \includegraphics[width=1\textwidth]{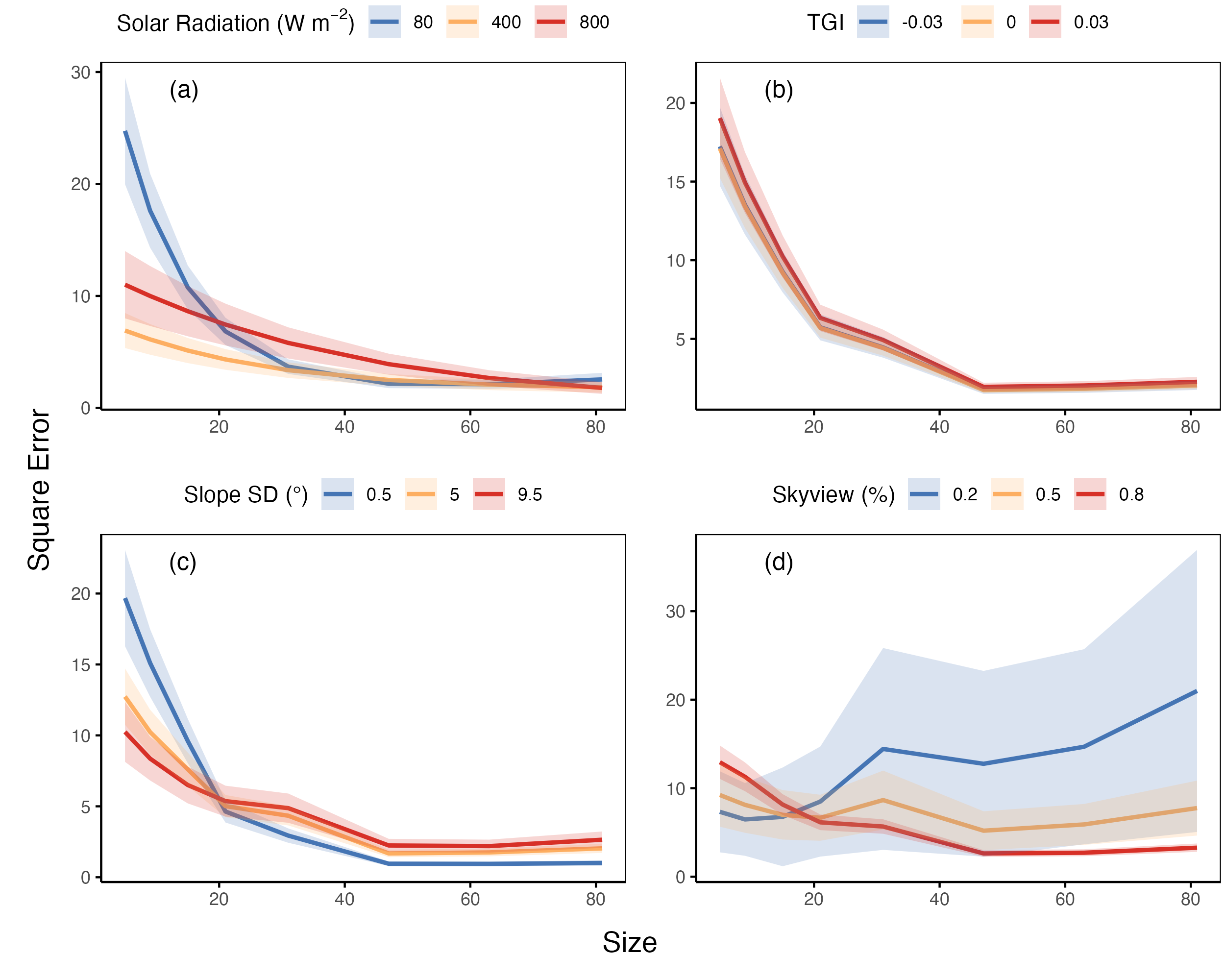}
    \caption{The improvement of model performance improved with increased spatial resolution varied with spatial characteristics. Each panel shows how tile size affects prediction accuracy (mean square error) under different (a) solar radiation, (b) vegetation cover (TGI), (c) changes in surface slopes (slope SD across 2m radius), and (d) skyview. Lines show smooth GAM fits with shaded confidence intervals. For each feature, we chose three levels that represent low (blue), intermediate (orange) and high (red) values in our maps}  \label{fig:tile_size_vs_spatial_features}
\end{figure}

\noindent Our statistical analysis of 20 pixels from each test map suggested that spatial features significantly affect how model performance is improved with tile size. In particular, model performance is improved with tiles size in pixels with low solar radiation (\textit{tile size} $\times$ \textit{solar radiation}: edf = 28.39, F = 133.071, \textit{p} < 0.001) (Fig. \ref{fig:tile_size_vs_spatial_features}a). Moreover, the rate of improvement with tile size was slightly but significantly higher in pixels with high vegetation cover (\textit{tile size}: edf = 6.225, F = 407.582, \textit{p} < 0.001; \textit{TGI}: edf = 6.307, F = 2.604, \textit{p} < 0.05) (Fig. \ref{fig:tile_size_vs_spatial_features}b), and low change in surface slope (\textit{tile size} $\times$ \textit{slope SD}: edf = 21.61, F = 132.610, \textit{p} < 0.001) (Fig. \ref{fig:tile_size_vs_spatial_features}c). Finally, our analysis showed that increasing tile size improved model predictions only in  pixels with high skyview (\textit{tile size} $\times$ \textit{skyview}: edf = 23.37, F = 156.03, \textit{p} < 0.001) (Fig. \ref{fig:tile_size_vs_spatial_features}d). 

\section{DISCUSSION}
\subsection{Deep learning as a tool to reveal the importance of spatial context in microclimate models}

\noindent
Despite major advances in physically based microclimate modeling \citep{briscoe_2023}, many existing frameworks estimate temperature independently for each spatial unit and rely on simplified representations of lateral heat exchange, making it difficult to isolate and quantify spatial interactions using traditional approaches. Here, we developed a deep learning–based method that addresses this gap by explicitly quantifying the contribution of spatial context to microclimate formation. Using this method, we show that ground temperature at a given location is strongly influenced by environmental conditions in surrounding microhabitats over spatial extents of several meters, with predictive gains saturating at approximately 5–7 m. This result indicates that microclimate temperatures are shaped not only by the surface properties of the modeled location, but also by horizontal heat transfer and radiative interactions operating across neighboring areas. Importantly, our deep learning model is general and transferable, allowing other researchers to quantify spatial dependencies in microclimate predictions across different ecosystems, spatial resolutions, and data sources.

By training repeatedly while systematically varying the spatial extent of the input data, our controlled manipulation of spatial context allows the contribution of nearby environmental information to be assessed directly. Our findings are consistent with known microclimate processes, including reflected shortwave radiation, longwave radiative exchange, and local modification of air temperatures by surrounding terrain and vegetation \citep{kearney_2017, rahman_2020}. Together, these results identify conditions under which assumptions of spatial independence commonly adopted in physically based microclimate models are likely to be insufficient, and where explicitly accounting for spatial interactions may improve model performance and realism.

\subsection{Spatial context varies across times of day and microclimate conditions}

\noindent
Our results demonstrate that the importance of spatial context in microclimate prediction is not constant, but varies systematically across times of day. In our desert habitat, the strongest improvements associated with increasing spatial extent were observed during morning and evening hours, whereas gains were more modest during midday (Fig. 4). These patterns are consistent with the underlying physical drivers of ground temperature. During morning and evening, low solar angles amplify the effects of shading from nearby objects, leading to strong lateral gradients in temperature. Rapid post-nighttime increases and post-daytime decreases in radiation during these periods can further amplify shade effects and the contribution of horizontal heat flux. Under these conditions, the temperature experienced at a given location is more strongly influenced by its surroundings, making spatially explicit information particularly informative. In contrast, during midday—when solar radiation is high and ground temperature is more strongly driven by local surface properties and direct vertical heating, reducing the relative contribution of nearby pixels.

Beyond temporal variation, we show that the importance of spatial information also depends on local microclimate characteristics. Model performance improved more strongly with increasing spatial context in areas of low solar radiation, low surface slope variability, and high sky view (Fig. 5). Under low solar radiation (such as during mornings, evenings, and shaded microhabitat), heat gain is less through radiation from the sun and therefore may be dominated by nearby pixels. Low slope variability and high skyview occurs in relatively flat and open habitats. It is possible that in terrestrial habitats, flat and homogeneous patches can support more coherent horizontal heat transport in the near-surface air than strongly heterogeneous patches, because the flow and turbulence are less disrupted and more spatially uniform \citep{li_2024}. In more structurally complex habitats, surface roughness and vegetation break up these interactions, causing temperature patterns to be dominated by very local conditions and reducing the influence of the broader surroundings \citep{li_2024}. Similarly, soil moisture gradients are often less uniform in heterogeneous patches due to localized run-on and run-off processes, potentially altering soil thermal conductivity and evaporative cooling, and thereby complicating vertical heat fluxes. Hence, assuming that microclimate conditions are determined solely by local pixel-level properties is likely to be insufficient in many settings. 

\subsection{Implications for physically based microclimate models}

\noindent
Our model highlights the importance of representation of horizontal heat transfer or spatial interactions among neighboring microhabitats in physical models. Such microclimate models are grounded in well-established physical principles and have proven highly valuable for ecological applications \citep{briscoe_2023}. However, as one of the current challenges is to find novel approaches to increase the accuracy of these models, our findings indicate that one simplifying assumption lies in simplifying horizontal heat transfer. Such simplifications may contribute to systematic prediction errors \citep{Itzkovitch2025}. The spatial scales identified here, on the order of several meters, suggest that lateral radiative and thermal interactions can meaningfully influence ground temperatures and should be considered when parameterizing or extending physical microclimate models.

Explicitly incorporating horizontal heat flow into microclimate models poses a substantial challenge for both model developers and users. Although lateral transport of heat and moisture is a core component of regional and global climate models \citep{taylor_2012, Skamarock2008}, implementing analogous processes at microclimate scales involves significant mathematical and computational difficulties. These include representing horizontal radiative exchange and near-surface heat advection at meter-scale resolution, as well as resolving fine-scale heterogeneity in surface properties and airflow. Historically, such complexity has limited the feasibility of spatially explicit microclimate models and motivated simplifying assumptions of spatial independence. In addition, parameterizing these processes requires high-resolution spatial data, which are often unavailable at large spatial extents or across long time periods. 

However, several recent developments are lowering these barriers. Advances in computational resources now allow increasingly complex simulations at fine spatial resolutions, while high-resolution environmental data are becoming more accessible through modern remote sensing technologies. In particular, the widespread availability of off-the-shelf drones enables detailed mapping of terrain, vegetation structure, and surface properties at spatial scales directly relevant to microclimate processes \citep{Itzkovitch2025}. In addition, repeated airborne remote sensing surveys, such as those conducted by the U.S. National Science Foundation’s National Ecological Observatory Network (NEON), provide standardized spatial data across multiple regions and time periods \citep{musinsky_2022}. Together, these developments make it increasingly realistic to incorporate spatial interactions into physical microclimate models.

Importantly, our results do not imply that physical models should be replaced by data-driven approaches. Rather, they highlight specific conditions under which existing formulations may be incomplete. By revealing when and where spatial context matters most, deep learning models can help identify gaps in physical representations and guide targeted model improvements. For example, incorporating simplified horizontal exchange terms, spatial smoothing informed by surrounding structure, or context-aware parameterization could improve predictions without sacrificing interpretability or generality.

\subsection{The potential of AI and hybrid modeling approaches}

\noindent
More broadly, our study illustrates the potential of artificial intelligence to complement mechanistic modeling in ecology. Deep neural networks are particularly well-suited for capturing nonlinear spatial dependencies and complex interactions among environmental variables, making them powerful tools for exploring fine-scale climate processes. When used thoughtfully, AI models can serve as hypothesis-generating instruments, revealing emergent spatial scales and context dependencies that may be difficult to anticipate a priori. 

Looking forward, hybrid modeling approaches that integrate physical microclimate models with machine learning offer a promising pathway for advancing microclimate prediction. Physical models provide mechanistic grounding and robustness beyond the range of observed data, while AI-based components can learn corrections, spatial interactions, or context-dependent effects directly from empirical observations \citep[e.g., ][]{wesselkamp_2024}. Such hybrid frameworks already improve predictive accuracy and reduce bias in global and regional climate models \citep[reviewed by ][]{levy_2024}. Our results provide a concrete basis for extending this paradigm to microclimate modeling by identifying the spatial scales over which spatial context becomes informative, thereby offering practical guidance for incorporating spatial interactions into physically grounded models such as NicheMapR and microclima. Finaly, AI models readily integrate heterogeneous data sources, including optical imagery, thermal imagery, and LiDAR, providing a flexible bridge between observational technologies, testing the importance of data for microclimate modeling, and eventually increase mechanistic understanding. 

Our study illustrates the potential of artificial intelligence to complement mechanistic modeling in ecology. Deep neural networks are particularly well-suited for capturing nonlinear spatial dependencies and complex interactions among environmental variables, making them powerful tools for exploring fine-scale climate processes. In addition, AI-based models are typically orders of magnitude faster to evaluate than full physical simulations, enabling large-scale inference, rapid sensitivity analyses, and scenario exploration that would otherwise be computationally prohibitive. When used thoughtfully, AI models can also serve as hypothesis-generating instruments, revealing emergent spatial scales and context dependencies that may be difficult to anticipate a priori.

Looking forward, hybrid modeling approaches that integrate physical microclimate models with machine learning offer a promising pathway for advancing microclimate prediction. Physical models provide mechanistic grounding and robustness beyond the range of observed data, while AI-based components can learn corrections, spatial interactions, or context-dependent effects directly from empirical observations \citep[e.g., ][]{wesselkamp_2024}. Importantly, the development of effective AI components benefits from close alignment with physical intuition: in our case, we explicitly assumed that dominant corrections arise from local neighborhood effects and therefore employed convolutional architectures. Such hybrid frameworks already improve predictive accuracy and reduce bias in global and regional climate models \citep[reviewed by ][]{levy_2024}. Our results provide a concrete basis for extending this paradigm to microclimate modeling by identifying the spatial scales over which spatial context becomes informative, thereby offering practical guidance for incorporating spatial interactions into physically grounded models such as NicheMapR and microclima.  

\subsection{Limitations and future directions}

\noindent
Despite the insights provided by our approach, several limitations should be acknowledged. First, our analysis focused on a single arid study system with relatively simple vegetation structure, limited vertical complexity, and limited moisture effects. While this setting is well-suited for isolating spatial effects on ground temperature, the spatial scales identified here may differ in ecosystems with denser canopies, stronger vertical stratification, or higher humidity, where turbulent heat exchange and evapotranspiration play a larger role. In addition, our models were trained and evaluated using drone-derived thermal imagery collected under specific seasonal and diurnal conditions. Although this design allowed us to capture a wide range of thermal conditions, extending the temporal scope—particularly to extreme weather events or transitional seasons—may reveal different spatial dependencies.

From a modeling perspective, our deep learning framework was intentionally designed to test the importance of spatial context rather than to provide a fully interpretable or mechanistic representation of microclimate processes. While variation in tile size offers indirect insight into the spatial scales over which nearby information influences predictions, it does not explicitly disentangle the relative contributions of individual physical processes such as shortwave reflection, longwave radiation exchange, or lateral heat transport through air and substrate. Future work could integrate explainable AI techniques \citep{oloughlin_2025} or process-informed neural networks \citep{wesselkamp_2024} to better link learned spatial patterns with specific heat-transfer mechanisms. In parallel, incorporating explicit horizontal exchange terms into physically based microclimate models, guided by the spatial scales identified here, may provide a promising pathway toward hybrid models that retain mechanistic transparency while benefiting from data-driven insights. For example, in structurally complex habitats such as forests, a deep neural network could estimate canopy-level temperature patterns and associated horizontal heat exchange across large vegetated areas. These predictions could then serve as inputs to physically based microclimate models that resolve ground-level energy balance and vertical heat transfer. Extending our framework across ecosystems and spatial scales will allow us to evaluate the generality of spatial coupling in microclimate dynamics and enhance predictions under ongoing climate change.

\section{CONCLUSIONS}
\noindent
We developed a deep neural network framework to quantify the importance of spatial context in microclimate temperature predictions. By systematically varying the spatial extent of model inputs, we showed that incorporating nearby environmental information substantially improves predictive accuracy, with gains saturating at spatial scales of several meters. The strength of these effects varied across times of day, microhabitats, and spatial characteristics, indicating that horizontal heat transfer and spatial heterogeneity play a context-dependent role in shaping ground temperatures. Rather than serving solely as a predictive tool, deep learning allowed us to identify when and where simplifying assumptions in traditional microclimate models are most likely to break down. Together, our findings highlight the value of integrating high-resolution spatial data with modeling approaches that explicitly account for spatial interactions, and they provide a foundation for developing hybrid mechanistic–machine learning microclimate models that are both physically grounded and ecologically relevant.

\section{ACKNOWLEDGMENTS} 
\noindent We thank Simon Jamison for logistical assistance. The research was funded by the National Geographic Society (NGS-84241T-21).

\section{SUPPLEMENTARY MATERIALS}

\subsection{Maps Development for Microclimate Modelling}
\noindent
Solar radiation was estimated as the total incoming shortwave radiation (direct and diffuse) reaching each pixel, accounting for local surface orientation. This layer was derived from the Digital Surface Model (DSM) by first calculating pixel-level slope and aspect using the \textit{r.slope.aspect} module in GRASS GIS. These terrain parameters were then used as inputs to the \textit{r.sun} module to compute spatially explicit solar radiation values.

The \textit{r.sun} calculations require information on solar geometry, including the day of year and solar time, which were obtained using the \textit{getSunlightTimes} function from the \textit{suncalc} R package \citep{Thieurmel_2022}. Because \textit{r.sun} estimates solar radiation under clear-sky conditions, modeled values were adjusted to account for cloud cover. This correction was applied by scaling the radiation map by the ratio between measured solar radiation recorded at the meteorological station and the maximum modeled radiation predicted for near-horizontal surfaces (slope < 0.1°).

\subsubsection*{Shade}
\noindent
Shade was represented as a binary spatial layer, with pixels classified as either shaded (1) or unshaded (0). This layer was derived using the \textit{r.sun} module in GRASS GIS by computing a solar incidence map, which describes the angle at which direct solar radiation reaches each pixel. Pixels assigned NULL values in the incidence output correspond to locations receiving no direct sunlight at the time of image acquisition and were therefore classified as shaded.

Shaded pixels in this analysis represent exposed ground surfaces that were temporarily shaded by nearby landscape elements, such as rocks, shrubs, or local topographic features, as determined by solar position during the survey. Areas beneath dense vegetation canopy were excluded from the dataset.

\subsubsection*{Skyview}
\noindent
Skyview was quantified as the skyview factor, defined as the proportion of visible sky at each pixel constrained by surrounding terrain and surface features such as rocks, shrubs, and trees. This layer was calculated using the \textit{r.skyview} module in GRASS GIS. In microclimate modeling, the skyview factor influences the balance of longwave radiation received at the surface, with lower values indicating greater obstruction by surrounding features and higher values corresponding to more open conditions. Skyview values range from 0 to 1, where values close to 0 indicate surfaces that are largely enclosed by surrounding features and values close to 1 indicate fully exposed locations with an unobstructed view of the sky.

\subsubsection*{Vegetation cover}
\noindent
Vegetation cover was quantified from the RGB orthophotos using the Triangular Greenness Index (TGI), calculated for each pixel as a proxy for chlorophyll content \citep{hunt_2013}.

\subsubsection*{Height}
\noindent
Height was quantified as the local elevation of surface features relative to the surrounding ground at each pixel. This layer was derived by calculating the difference between the Digital Surface Model (DSM), which captures the elevation of surface features such as rocks, terraces, and low vegetation, and the Digital Terrain Model (DTM), which represents the underlying bare ground surface. The resulting difference provides an estimate of object height and microtopographic variation across the landscape \citep[following][]{duffy_2021}. Values approach zero in open areas and increase where surface features protrude above the ground.

\newpage

\begin{table}
\centering
\begin{tabular}{ |p{2cm}||p{1.5cm}|p{5.5cm}|p{6.5cm}| }
 \hline
 Symbol & Units & Description & Equation / Source\\
 \hline
 $\alpha$ & dec. \% & Albedo & GLDAS \\
 \hline
 \textit{SHD} & 0 - No \newline 1 - Yes & The presence of shade & \textit{r.sun} GRASS command  \\
 \hline
 $S_{g}$ & Wm\textsuperscript{-2} & Solar radiation that reaches the ground & Station  \\
 \hline
 \textit{SVF} & dec. \% & Skyview & \textit{r.skyview} GRASS command  \\
 \hline
 $T_g$ & K & Surface temperature. Initial condition only. & Station \\
 \hline
 $T_a$ & K & Air temperature at 1.2m height & Station \\
 \hline
 \textit{P} & Pascal & Ground pressure  & Station \\
 \hline
 \textit{STEMP} & K & Soil temperature in four soil layers: 10, 40, 100, 200 cm & GLDAS \\
 \hline
 \textit{SMOIS} & m\textsuperscript{3} m\textsuperscript{-3} & Soil moisture in four soil layers: 10, 40, 100, 200 cm & GLDAS \\
 \hline
 \textit{RH} & \% & Relative humidity & Station \\
 \hline
 \end{tabular}
\caption{Input features used in the modified ResNet model.}
\label{table:physparam}
\end{table}

\newpage

\begin{table}
\centering
\begin{tabular}{|l|l|l|}
\hline
Set & Date & Time \\ \hline
Train & 18.9.19 & 09:00, 12:00, 14:00, 15:00, 17:20 \\ \hline
 & 29.5.19 & 08:30, 16:50, 17:30 \\ \hline
 & 30.1.20 & 09:20, 09:50, 10:50, 12:00, 13:00, 13:50, 14:49, 15:23 \\ \hline
 & 30.5.19 & 06:00, 06:30 \\ \hline
 & 7.11.19 & 10:30, 11:00, 13:10, 15:50, 16:40 \\ \hline
Test & 12.04.21 & 11:18 \\ \hline
 & 23.09.19 & 06:10, 07:00, 08:00, 14:10, 15:10, 16:10 \\ \hline
 & 31.05.21 & 15:16, 17:12, 18:05 \\ \hline
\end{tabular}
\caption{Dates and times of the drone flight surveys. Flights conducted on different dates were used for training and testing the bias-correction machine learning model.}
\label{table:flights}
\end{table}

\clearpage

\newpage

\bibliography{references}

@article{kearney_2017,
title = {{NicheMapR} - an {R} package for biophysical modelling: the microclimate model},
author = {Kearney, Michael R. and Porter, Warren P.},
pages = {664-674},
url = {http://doi.wiley.com/10.1111/ecog.02360},
year = {2017},
month = {oct},
urldate = {2020-10-19},
journal = {Ecography},
volume = {40},
issn = {09067590},
doi = {10.1111/ecog.02360},
sciwheel-projects = {insects\_and\_microclimate}
}

@article{maclean_2019,
title = {Microclima: An R package for modelling meso- and microclimate},
author = {Maclean, Ilya M. D. and Mosedale, Jonathan R. and Bennie, Jonathan J.},
pages = {280-290},
url = {https://onlinelibrary.wiley.com/doi/abs/10.1111/2041-{210X}.13093},
year = {2019},
month = {feb},
urldate = {2020-08-31},
journal = {Methods in Ecology and Evolution},
volume = {10},
number = {2},
issn = {2041-{210X}},
doi = {10.1111/2041-{210X}.13093},
sciwheel-projects = {insects\_and\_microclimate}
}

@article{kearney_2020,
title = {A method for computing hourly, historical, terrain-corrected microclimate anywhere on earth},
author = {Kearney, Michael R. and Gillingham, Phillipa K. and Bramer, Isobel and Duffy, James P. and Maclean, Ilya M.D.},
pages = {38-43},
url = {https://onlinelibrary.wiley.com/doi/abs/10.1111/2041-{210X}.13330},
year = {2020},
month = {jan},
urldate = {2020-08-31},
journal = {Methods in Ecology and Evolution},
volume = {11},
number = {1},
issn = {2041-{210X}},
doi = {10.1111/2041-{210X}.13330},
sciwheel-projects = {insects\_and\_microclimate}
}

@article{tabak_2018,
title = {Machine learning to classify animal species in camera trap images: Applications in ecology},
author = {Tabak, Michael A. and Norouzzadeh, Mohammad S. and Wolfson, David W. and Sweeney, Steven J. and Vercauteren, Kurt C. and Snow, Nathan P. and Halseth, Joseph M. and Di Salvo, Paul A. and Lewis, Jesse S. and White, Michael D. and Teton, Ben and Beasley, James C. and Schlichting, Peter E. and Boughton, Raoul K. and Wight, Bethany and Newkirk, Eric S. and Ivan, Jacob S. and Odell, Eric A. and Brook, Ryan K. and Lukacs, Paul M. and Moeller, Anna K. and Mandeville, Elizabeth G. and Clune, Jeff and Miller, Ryan S.},
pages = {585-590},
url = {http://doi.wiley.com/10.1111/2041-{210X}.13120},
year = {2018},
month = {nov},
day = {26},
urldate = {2020-09-09},
journal = {Methods in Ecology and Evolution},
volume = {10},
number = {4},
issn = {2041210X},
doi = {10.1111/2041-{210X}.13120},
sciwheel-projects = {insects\_and\_microclimate}
}

@article{levy_2016a,
title = {Foraging activity pattern is shaped by water loss rates in a diurnal desert rodent.},
author = {Levy, Ofir and Dayan, Tamar and Porter, Warren P and Kronfeld-Schor, Noga},
pages = {205-218},
url = {http://dx.doi.org/10.1086/687246},
year = {2016},
month = {aug},
urldate = {2019-08-21},
journal = {The American Naturalist},
volume = {188},
number = {2},
doi = {10.1086/687246},
pmid = {27420785},
sciwheel-projects = {wild\_boars}
}

@article{potter_2013,
title = {Microclimatic challenges in global change biology.},
author = {Potter, Kristen A and Arthur Woods, H and Pincebourde, Sylvain},
pages = {2932-2939},
url = {http://dx.doi.org/10.1111/gcb.12257},
year = {2013},
month = {oct},
urldate = {2019-09-05},
journal = {Global Change Biology},
volume = {19},
number = {10},
doi = {10.1111/gcb.12257},
pmid = {23681970}
}

@article{pincebourde_2020,
title = {There is plenty of room at the bottom: microclimates drive insect vulnerability to climate change.},
author = {Pincebourde, Sylvain and Woods, H Arthur},
pages = {63-70},
url = {https://linkinghub.elsevier.com/retrieve/pii/S2214574520300870},
year = {2020},
month = {jul},
day = {16},
urldate = {2020-08-16},
journal = {Current opinion in insect science},
volume = {41},
issn = {22145745},
doi = {10.1016/j.cois.2020.07.001},
pmid = {32777713},
sciwheel-projects = {urban ecology}
}

@article { TheGlobalLandDataAssimilationSystem,
      author = "M. Rodell and P. R. Houser and U. Jambor and J. Gottschalck and K. Mitchell and C.-J. Meng and K. Arsenault and B. Cosgrove and J. Radakovich and M. Bosilovich and J. K. Entin and J. P. Walker and D. Lohmann and D. Toll",
      title = "The Global Land Data Assimilation System",
      journal = "Bulletin of the American Meteorological Society",
      year = "2004",
      publisher = "American Meteorological Society",
      address = "Boston MA, USA",
      volume = "85",
      number = "3",
      doi = "https://doi.org/10.1175/BAMS-85-3-381",
      pages=      "381 - 394",
      url = "https://journals.ametsoc.org/view/journals/bams/85/3/bams-85-3-381.xml"
}

@article{BIVAND20001043,
title = {Using the {R} statistical data analysis language on GRASS 5.0 GIS database files},
journal = {Computers and Geosciences},
volume = {26},
number = {9},
pages = {1043-1052},
year = {2000},
issn = {0098-3004},
doi = {https://doi.org/10.1016/S0098-3004(00)00057-1},
url = {https://www.sciencedirect.com/science/article/pii/S0098300400000571},
author = {Roger S. Bivand}
}

@article{briscoe_2023,
title = {Mechanistic forecasts of species responses to climate change: The promise of biophysical ecology.},
author = {Briscoe, Natalie J and Morris, Shane D and Mathewson, Paul D and Buckley, Lauren B and Jusup, Marko and Levy, Ofir and Maclean, Ilya M D and Pincebourde, Sylvain and Riddell, Eric A and Roberts, Jessica A and Schouten, Rafael and Sears, Michael W and Kearney, Michael Ray},
pages = {1451-1470},
url = {https://onlinelibrary.wiley.com/doi/10.1111/gcb.16557},
year = {2023},
month = {mar},
urldate = {2023-03-08},
journal = {Global Change Biology},
volume = {29},
number = {6},
issn = {1354-1013},
doi = {10.1111/gcb.16557},
pmid = {36515542},
sciwheel-projects = {Stark\_et\_al\_ICB}
}

@article{moncaz_2012,
title = {Breeding sites of Phlebotomus sergenti, the sand fly vector of cutaneous leishmaniasis in the Judean Desert.},
author = {Moncaz, Aviad and Faiman, Roy and Kirstein, Oscar and Warburg, Alon},
pages = {e1725},
url = {http://dx.doi.org/10.1371/journal.pntd.0001725},
year = {2012},
month = {jul},
day = {3},
urldate = {2021-12-22},
journal = {{PLoS} Neglected Tropical Diseases},
volume = {6},
number = {7},
doi = {10.1371/journal.pntd.0001725},
pmid = {22802981},
pmcid = {PMC3389037},
sciwheel-projects = {Stark\_et\_al\_ICB}
}

@Manual{GRASS_GIS_software,
    title = {Geographic Resources Analysis Support System (GRASS GIS) Software, Version 7.4},
    author = {{GRASS Development Team}},
    organization = {Open Source Geospatial Foundation},
    year = {2022},
    url = {https://grass.osgeo.org},
    doi = {https://doi.org/10.5281/zenodo.5176030}
  }

@article{levy_2024,
    author = {Levy, Ofir and Shahar, Shimon},
    title = "{Artificial Intelligence for Climate Change Biology: From Data Collection to Predictions}",
    journal = {Integrative and Comparative Biology},
    pages = {icae127},
    year = {2024},
    month = {07},
    issn = {1540-7063},
    doi = {10.1093/icb/icae127},
    url = {https://doi.org/10.1093/icb/icae127},
    eprint = {https://academic.oup.com/icb/advance-article-pdf/doi/10.1093/icb/icae127/58886050/icae127.pdf},
}

@book{burnham2002model,
  author = {Burnham, K.P. and Anderson, D.R.},
  publisher = {Springer Verlag},
  title = {Model selection and multimodel inference: a practical information-theoretic approach},
  username = {jabreftest},
  year = 2002
}

@article{liang_2020,
title = {Image classification based on {RESNET}},
author = {Liang, Jiazhi},
pages = {012110},
url = {https://iopscience.iop.org/article/10.1088/1742-6596/1634/1/012110},
year = {2020},
month = {sep},
urldate = {2021-10-05},
journal = {Journal of Physics: Conference Series},
volume = {1634},
issn = {1742-6588},
doi = {10.1088/1742-6596/1634/1/012110}
}

@article{schneider_2023,
title = {Harnessing {AI} and computing to advance climate modelling and prediction},
author = {Schneider, Tapio and Behera, Swadhin and Boccaletti, Giulio and Deser, Clara and Emanuel, Kerry and Ferrari, Raffaele and Leung, L. Ruby and Lin, Ning and Müller, Thomas and Navarra, Antonio and Ndiaye, Ousmane and Stuart, Andrew and Tribbia, Joseph and Yamagata, Toshio},
pages = {887-889},
url = {https://www.nature.com/articles/s41558-023-01769-3},
year = {2023},
month = {sep},
urldate = {2023-09-06},
journal = {Nature Climate Change},
volume = {13},
number = {9},
issn = {1758-{678X}},
doi = {10.1038/s41558-023-01769-3},
sciwheel-projects = {microclimate\_model}
}

@article{star_2020,
title = {Comparing {RGB} - based vegetation indices from {UAV} imageries to estimate hops canopy area},
author = {Starý, K. and Jelínek, Z. and Kumhálová, J. and Chyba, J. and Balážová, K.},
url = {https://dspace.emu.ee/xmlui/handle/10492/6091},
year = {2020},
urldate = {2024-11-02},
journal = {Agronomy Research},
doi = {10.15159/ar.20.169},
sciwheel-projects = {microclimate\_model}
}

@Manual{R_Core_Team,
  title = {R: A Language and Environment for Statistical Computing},
  author = {{R Core Team}},
  organization = {R Foundation for Statistical Computing},
  address = {Vienna, Austria},
  year = {2021},
  url = {https://www.R-project.org/},
}

@Manual{Thieurmel_2022,
    title = {suncalc: Compute Sun Position, Sunlight Phases, Moon
      Position and Lunar Phase},
    author = {Benoit Thieurmel and Achraf Elmarhraoui},
    year = {2022},
    note = {R package version 0.5.1},
    url = {https://github.com/datastorm-open/suncalc},
  }

@article{rahman_2020,
title = {Tree cooling effects and human thermal comfort under contrasting species and sites},
author = {Rahman, Mohammad A. and Hartmann, Christian and Moser-Reischl, Astrid and von Strachwitz, Miriam Freifrau and Paeth, Heiko and Pretzsch, Hans and Pauleit, Stephan and Rötzer, Thomas},
pages = {107947},
url = {https://linkinghub.elsevier.com/retrieve/pii/S0168192320300496},
year = {2020},
month = {jun},
urldate = {2023-09-11},
journal = {Agricultural and Forest Meteorology},
volume = {287},
issn = {01681923},
doi = {10.1016/j.agrformet.2020.107947},
sciwheel-projects = {Zlotnick\_et\_al\_PNAS}
}

@article{duffy_2021,
title = {Drones provide spatial and volumetric data to deliver new insights into microclimate modelling},
author = {Duffy, James P. and Anderson, Karen and Fawcett, Dominic and Curtis, Robin J. and Maclean, Ilya M. D.},
pages = {685-702},
url = {http://link.springer.com/10.1007/s10980-020-01180-9},
year = {2021},
month = {mar},
urldate = {2024-02-17},
journal = {Landscape Ecology},
volume = {36},
number = {3},
issn = {0921-2973},
doi = {10.1007/s10980-020-01180-9},
sciwheel-projects = {Shermeister\_MEE}
}

@article{campsvalls_2025,
title = {Artificial intelligence for modeling and understanding extreme weather and climate events.},
author = {Camps-Valls, Gustau and Fernández-Torres, Miguel-Angel and Cohrs, Kai-Hendrik and Höhl, Adrian and Castelletti, Andrea and Pacal, Aytac and Robin, Claire and Martinuzzi, Francesco and Papoutsis, Ioannis and Prapas, Ioannis and Pérez-Aracil, Jorge and Weigel, Katja and Gonzalez-Calabuig, Maria and Reichstein, Markus and Rabel, Martin and Giuliani, Matteo and Mahecha, Miguel D and Popescu, Oana-Iuliana and Pellicer-Valero, Oscar J and Ouala, Said and Salcedo-Sanz, Sancho and Sippel, Sebastian and Kondylatos, Spyros and Happé, Tamara and Williams, Tristan},
pages = {1919},
url = {http://dx.doi.org/10.1038/s41467-025-56573-8},
year = {2025},
month = {feb},
day = {24},
urldate = {2025-03-09},
journal = {Nature Communications},
volume = {16},
number = {1},
doi = {10.1038/s41467-025-56573-8},
pmid = {39994190},
pmcid = {PMC11850610},
sciwheel-projects = {microclimate\_model}
}

@article{slater_2023,
title = {Hybrid forecasting: blending climate predictions with {AI} models},
author = {Slater, Louise J. and Arnal, Louise and Boucher, Marie-Amélie and Chang, Annie Y.-Y. and Moulds, Simon and Murphy, Conor and Nearing, Grey and Shalev, Guy and Shen, Chaopeng and Speight, Linda and Villarini, Gabriele and Wilby, Robert L. and Wood, Andrew and Zappa, Massimiliano},
pages = {1865-1889},
url = {https://hess.copernicus.org/articles/27/1865/2023/},
year = {2023},
month = {may},
day = {15},
urldate = {2025-03-09},
journal = {Hydrology and Earth System Sciences},
volume = {27},
number = {9},
issn = {1607-7938},
doi = {10.5194/hess-27-1865-2023},
sciwheel-projects = {microclimate\_model}
}

@article{lecun_2015,
title = {Deep learning},
author = {{LeCun}, Y and Bengio, Y and Hinton, G},
pages = {436-444},
url = {http://www.nature.com/doifinder/10.1038/nature14539},
year = {2015},
month = {may},
day = {28},
urldate = {2018-06-27},
journal = {Nature},
volume = {521},
number = {7553},
issn = {0028-0836},
doi = {10.1038/nature14539},
pmid = {26017442},
sciwheel-projects = {Levy\_ICB\_2024}
}

@Article{mgcv_2011,
    title = {Smoothing parameter and model selection for general smooth models (with discussion)},
    author = {S.N. Wood and {N.} and {Pya} and B. S{"a}fken},
    journal = {Journal of the American Statistical Association},
    year = {2016},
    pages = {1548-1575},
    volume = {111},
  }

@incollection{NEURIPS2019_9015,
title = {PyTorch: An Imperative Style, High-Performance Deep Learning Library},
author = {Paszke, Adam and Gross, Sam and Massa, Francisco and Lerer, Adam and Bradbury, James and Chanan, Gregory and Killeen, Trevor and Lin, Zeming and Gimelshein, Natalia and Antiga, Luca and Desmaison, Alban and Kopf, Andreas and Yang, Edward and DeVito, Zachary and Raison, Martin and Tejani, Alykhan and Chilamkurthy, Sasank and Steiner, Benoit and Fang, Lu and Bai, Junjie and Chintala, Soumith},
booktitle = {Advances in Neural Information Processing Systems 32},
pages = {8024--8035},
year = {2019},
publisher = {Curran Associates, Inc.},
url = {http://papers.neurips.cc/paper/9015-pytorch-an-imperative-style-high-performance-deep-learning-library.pdf}
}

@Manual{terra_2025,
    title = {terra: Spatial Data Analysis},
    author = {Robert J. Hijmans},
    year = {2025},
    note = {R package version 1.8-54},
    url = {https://CRAN.R-project.org/package=terra},
  }

@article {Itzkovitch2025,
	author = {Itzkovitch, Alon and Sulami, Idan and Efroni, Ronny Doron and Shahar, Moni and Levy, Ofir},
	title = {From Big Data to Small Scales: Machine Learning Enhances Microclimate Model Predictions},
	elocation-id = {2025.12.01.691551},
	year = {2025},
	doi = {10.64898/2025.12.01.691551},
	publisher = {Cold Spring Harbor Laboratory},
	URL = {https://www.biorxiv.org/content/early/2025/12/02/2025.12.01.691551},
	eprint = {https://www.biorxiv.org/content/early/2025/12/02/2025.12.01.691551.full.pdf},
	journal = {bioRxiv}
}

@article{li_2024,
title = {The effect of heterogeneous geometry on steady-state heat transfer in extrusion-based {3D} printed structures},
author = {Li, Zhengrong and Xing, Wenjing and Wang, Heyu and Sun, Jingting},
pages = {111147},
url = {https://linkinghub.elsevier.com/retrieve/pii/S2352710224027153},
year = {2024},
month = {dec},
urldate = {2026-01-19},
journal = {Journal of Building Engineering},
volume = {98},
issn = {23527102},
doi = {10.1016/j.jobe.2024.111147},
sciwheel-projects = {microclimate\_model}
}

@article{taylor_2012,
title = {An Overview of {CMIP5} and the Experiment Design},
author = {Taylor, Karl E. and Stouffer, Ronald J. and Meehl, Gerald A.},
pages = {485-498},
url = {http://journals.ametsoc.org/doi/abs/10.1175/{BAMS}-D-11-00094.1},
year = {2012},
month = {apr},
urldate = {2014-05-10},
journal = {Bulletin of the American Meteorological Society},
volume = {93},
number = {4},
issn = {0003-0007},
doi = {10.1175/{BAMS}-D-11-00094.1}
}

@techreport{Skamarock2008,
  author = {Skamarock, William C. and Klemp, Joseph B. and Dudhia, Jimy and Gill, Daniel and Barker, Michael and Duda, Jan and Huang, Xiang-Yu and Wang, Wei-Kuo and Zhang, Da-Lin},
  title = {A Description of the Advanced Research WRF Model Version 3},
  institution = {National Center for Atmospheric Research (NCAR)},
  type = {Technical Note},
  number = {NCAR/TN-475+STR},
  year = {2008},
  month = {June},
  url = {https://opensky.ucar.edu/system/files/2024-08/technotes_500.pdf},
  note = {The official citation reference for WRF Version 3 [1, 9, 10].}
}

@article{musinsky_2022,
title = {Spanning scales: The airborne spatial and temporal sampling design of the National Ecological Observatory Network},
author = {Musinsky, John and Goulden, Tristan and Wirth, Gregory and Leisso, Nathan and Krause, Keith and Haynes, Mitch and Chapman, Cameron},
pages = {1866-1884},
url = {https://onlinelibrary.wiley.com/doi/10.1111/2041-{210X}.13942},
year = {2022},
month = {sep},
urldate = {2026-01-22},
journal = {Methods in Ecology and Evolution},
volume = {13},
number = {9},
issn = {2041-{210X}},
doi = {10.1111/2041-{210X}.13942}
}

@article{wesselkamp_2024,
title = {Process-Informed Neural Networks: A Hybrid Modelling Approach to Improve Predictive Performance and Inference of Neural Networks in Ecology and Beyond.},
author = {Wesselkamp, Marieke and Moser, Niklas and Kalweit, Maria and Boedecker, Joschka and Dormann, Carsten F},
pages = {e70012},
url = {http://dx.doi.org/10.1111/ele.70012},
year = {2024},
month = {nov},
urldate = {2026-01-22},
journal = {Ecology Letters},
volume = {27},
number = {11},
doi = {10.1111/ele.70012},
pmid = {39625058},
pmcid = {PMC11613309},
sciwheel-projects = {microclimate\_model}
}

@article{oloughlin_2025,
title = {Moving beyond post hoc explainable artificial intelligence: a perspective paper on lessons learned from dynamical climate modeling},
author = {O'Loughlin, Ryan J. and Li, Dan and Neale, Richard and O'Brien, Travis A.},
pages = {787-802},
url = {https://gmd.copernicus.org/articles/18/787/2025/},
year = {2025},
month = {feb},
day = {11},
urldate = {2026-01-22},
journal = {Geoscientific Model Development},
volume = {18},
number = {3},
issn = {1991-9603},
doi = {10.5194/gmd-18-787-2025},
sciwheel-projects = {microclimate\_model}
}

@article{hunt_2013,
title = {A visible band index for remote sensing leaf chlorophyll content at the canopy scale},
author = {Hunt, E. Raymond and Doraiswamy, Paul C. and {McMurtrey}, James E. and Daughtry, Craig S.T. and Perry, Eileen M. and Akhmedov, Bakhyt},
pages = {103-112},
url = {https://linkinghub.elsevier.com/retrieve/pii/S0303243412001791},
year = {2013},
month = {apr},
urldate = {2020-10-12},
journal = {International Journal of Applied Earth Observation and Geoinformation},
volume = {21},
issn = {03032434},
doi = {10.1016/j.jag.2012.07.020},
sciwheel-projects = {Stark\_et\_al\_ICB}
}

@inproceedings{kingma2015adam,
  author    = {Diederik P. Kingma and Jimmy Ba},
  title     = {Adam: A Method for Stochastic Optimization},
  booktitle = {International Conference on Learning Representations (ICLR)},
  year      = {2015},
  url       = {https://arxiv.org/abs/1412.6980}
}

\end{document}